\documentclass[lettersize,journal]{IEEEtran}
\usepackage{amsmath,amsfonts}
\usepackage{algorithmic}
\usepackage{algorithm}
\usepackage{array}
\usepackage[caption=false,font=normalsize,labelfont=sf,textfont=sf]{subfig}
\usepackage{textcomp}
\usepackage{stfloats}
\usepackage{url}
\usepackage{verbatim}
\usepackage{graphicx}
\usepackage{amsmath}
\usepackage{color,soul}
\usepackage[dvipsnames]{xcolor}
\usepackage[numbers, sort&compress]{natbib}
\usepackage{enumitem}
\setlist{leftmargin=8pt}
\setlist[itemize]{topsep=0pt, itemsep=0pt, parsep=0pt, labelsep=2pt}

\usepackage[hidelinks]{hyperref}
\usepackage{tikz}

\usepackage[framemethod=TikZ]{mdframed}
\usepackage{lipsum}
\definecolor{hlcolor}{RGB}{255,231,173}
\sethlcolor{hlcolor}

\mdfdefinestyle{LLMFrame}{%
    linecolor=blue,
    outerlinewidth=1pt,
    roundcorner=10pt,
    innertopmargin=4pt,
    innerbottommargin=4pt,
    innerrightmargin=5pt,
    innerleftmargin=5pt,
    backgroundcolor=white}
    
\usepackage{balance}
\newcommand{\RN}[1]{%
  \textup{\uppercase\expandafter{\romannumeral#1}}%
}

\definecolor{color1}{RGB}{97,127,178}

\ifCLASSINFOpdf
\else
\fi

\begin{document}
\title{V2X-LLM: Enhancing V2X Integration and Understanding in Connected Vehicle Corridors}

\author{Keshu Wu, Pei Li, Yang Zhou*, Rui Gan, Junwei You, Yang Cheng, Jingwen Zhu, Steven T. Parker, Bin Ran, David A. Noyce, Zhengzhong Tu% <-this % stops a space
\thanks{*Corresponding author. (Email: yangzhou295@tamu.edu).}
\thanks{K. Wu was previously with the Department of Civil and Environmental Engineering at the University of Wisconsin-Madison, Madison, WI, 53706, USA. He is currently affiliated with the Department of Landscape Architecture and Urban Planning, and the Zachry Department of Civil and Environmental Engineering, Texas A\&M University, College Station, TX, 77840, USA.} %(email: keshuw@tamu.edu)
\thanks{Y. Zhou is associated with the Zachry Department of Civil and Environmental Engineering, Texas A\&M University, College Station, TX, 77840, USA.} % (email: yangzhou295@tamu.edu)
\thanks{P. Li, R. Gan, J. You, C. Yang, J. Zhu, S. Parker, B. Ran, and D. Noyce are all associated with the Department of Civil and Environmental Engineering at the University of Wisconsin-Madison, Madison, WI, 53706, USA.} % (e-mail: pei.li@wisc.edu, rgan6@wisc.edu, jyou38@wisc.edu, cheng8@wisc.edu, jzhu254@wisc.edu, sparker@engr.wisc.edu, bran@wisc.edu, and danoyce@wisc.edu)
\thanks{Z. Tu is associated with the Department of Computer Science and Engineering, Texas A\&M University, College Station, TX, 77843, USA.} %(email: tzz@tamu.edu)
}

\maketitle

\begin{abstract}
The advancement of Connected and Automated Vehicles (CAVs) and Vehicle-to-Everything (V2X) offers significant potential for enhancing transportation safety, mobility, and sustainability. However, the integration and analysis of the diverse and voluminous V2X data, including Basic Safety Messages (BSMs) and Signal Phase and Timing (SPaT) data, present substantial challenges, especially on Connected Vehicle Corridors. These challenges include managing large data volumes, ensuring real-time data integration, and understanding complex traffic scenarios. Although these projects have developed an advanced CAV data pipeline that enables real-time communication between vehicles, infrastructure, and other road users for managing connected vehicle and roadside unit (RSU) data, significant hurdles in data comprehension and real-time scenario analysis and reasoning persist. To address these issues, we introduce the V2X-LLM framework, a novel enhancement to the existing CV data pipeline. V2X-LLM leverages Large Language Models (LLMs) to improve the understanding and real-time analysis of V2X data. The framework includes four key tasks: Scenario Explanation, offering detailed narratives of traffic conditions; V2X Data Description, detailing vehicle and infrastructure statuses; State Prediction, forecasting future traffic states; and Navigation Advisory, providing optimized routing instructions. By integrating LLM-driven reasoning with V2X data within the data pipeline, the V2X-LLM framework offers real-time feedback and decision support for traffic management. This integration enhances the accuracy of traffic analysis, safety, and traffic optimization. Demonstrations in a real-world urban corridor highlight the framework's potential to advance intelligent transportation systems.

\end{abstract}

\begin{IEEEkeywords}
Connected vehicle corridor, large language models, data pipeline, intelligent transportation systems.
\end{IEEEkeywords}

\section{Introduction}
\label{sec:intro}

The emergence of Connected and Automated Vehicles (CAVs) and Vehicle-to-Everything (V2X)~\cite{yu2023v2x} presents a promising advancement in enhancing safety, increasing mobility, and promoting environmental sustainability in modern transportation systems~\cite{ran2019development, Joshua2017}. This global transition towards heightened interconnectivity and automation is integral to the evolution of intelligent transportation systems, particularly through the establishment of connected vehicle corridors~\cite{saroj2021development}. These corridors create an interconnected ecosystem where vehicles, infrastructure, and other road users communicate, facilitating the exchange of crucial information to enhance safety, efficiency, and mobility.

A growing number of research institutions, in collaboration with municipal and state transportation agencies, are developing Smart Corridors to facilitate CAV applications~\cite{wu2023development, li2023does, chen2023proactive}. These corridors are equipped with Roadside Units (RSUs) along urban arterials, providing a real-world testbed to evaluate advanced CAV technologies. Such initiatives enhance traffic safety, improve mobility, and advance CAV readiness by enabling robust data collection and analysis frameworks, exampled by~\cite{chowduri2024design, maglogiannis2021experimental, caruntu2021cyber}. These Connected Vehicle Corridor projects include a dedicated data pipeline and computational infrastructure to manage large-scale data generated from connected vehicle interactions. The corresponding data pipeline systems, are usually supported by high-performance computing resources to ensure seamless data acquisition, processing, and archiving, which lays the foundation to fulfill advanced CAV control such as~\cite{yao2021lane, guo2019joint, lu2021optimization}.

% \begin{figure}[!t]
%   \centering
%   \includegraphics[width=0.45\textwidth]{figs/deployment.png}
%   \caption{Connected Vehicle Corridor Deployment. (a) The birds-eye view of an intersection deployed with RSU. (b) The data transmission among RSU, signal controller, and OBU. (c) The layout of the connected vehicle corridor.}\label{fig:deploy}
% \end{figure}

% \begin{figure*}[!ht]
%   \centering
%   \includegraphics[width=0.8\textwidth]{figs/data_pipeline.png}
%   \caption{Connected Vehicle Corridor Data Pipeline}\label{fig:data_pipeline}
% \end{figure*}

Despite the effort by ~\cite{wu2023development, li2023does, wu2024digital}, significant challenges remain in optimizing the data pipeline for connected vehicle corridors. Managing the vast amounts of data generated by numerous connected vehicles and infrastructures requires advanced storage, processing, and retrieval solutions. The high frequency of V2X data, such as Signal Phase and Timing (SPaT), Intersection Mapping (MAP) data from RSUs, and Basic Safety Messages (BSMs) from On-Board Units (OBUs)~\cite{sae-j2735_2016}, necessitates real-time processing capabilities to maintain accurate and timely information flow~\cite{WAGNER2023}. Additionally, the diversity of data types—from structured packets to complex unstructured sensor readings—complicates the integration and analysis processes, making it challenging to extract actionable insights. Beyond multimodality, these data streams exhibit heterogeneity, spatiotemporal dependencies, and strict latency constraints. They combine discrete event messages with continuous sensor data, requiring synchronization across time and space. The real-time nature of CAV applications also demands low-latency processing to ensure safety-critical operations~\cite{yu2023v2x, aissaoui2014advanced, zhao2022collaborative}. Another challenge is understanding and reasoning about complex traffic scenarios. These include interpreting the context of traffic events, predicting vehicle behaviors, and managing dynamic changes in traffic conditions~\cite{zhang2023study}. Traditional approaches often fall short in providing comprehensive situational awareness and real-time decision-making support, especially when dealing with the vast and diverse datasets typical of V2X systems. Previous solutions have mainly focused on optimizing data transmission protocols, employing cloud or fog computing for enhanced data processing, and utilizing sophisticated data analytics algorithms~\cite{saroj2021development, feng2019}. However, these methods often lack the advanced interpretative capabilities needed to fully understand and respond to dynamic traffic situations.

LLMs, built on transformer architectures, are highly effective in processing sequential data, capturing long-range dependencies, and adapting to contextual variations dynamically\cite{floridi2020gpt, achiam2023gpt, touvron2023llama, ouyang2022training}. Their self-attention mechanism allows efficient handling of multimodal data, making them well-suited for interpreting diverse information, reasoning through complex patterns, and generating structured insights. In the field of intelligent transportation, LLMs introduce a scalable and adaptive approach to processing large-scale traffic data, understanding vehicle-infrastructure interactions, and improving situational awareness\cite{zheng2024large, peng2024lc, lai2023large, tang2024large, movahedi2024crossroads}. Their ability to interpret structured and unstructured data enables vehicle coordination, traffic flow optimization, and predictive analytics, which are crucial for connected and automated environments~\cite{azarafza2024hybrid, li2022v2x}. Specifically, in V2X communication systems, where heterogeneous, high-frequency data streams must be processed with low latency, LLMs can enhance data integration, real-time decision support, and predictive modeling to improve traffic safety and efficiency~\cite{wang2024llm, zhang2024trafficgpt, mahmud2025integrating}. By leveraging their capability to extract contextual insights from diverse data sources, LLMs can be seamlessly integrated into V2X data pipelines to support high-performance computation without compromising performance~\cite{wei2022chain}. LLMs have demonstrated strong reasoning abilities, allowing them to explain predictions, infer latent relationships in traffic data, and enhance interpretability in V2X-driven decision-making~\cite{rajani2019explain, huang2023can}.

In response to these challenges, we propose V2X-LLM, the first attempt to leverage LLMs for enhancing the integration and understanding of V2X smart corridors. While previous research has explored AI-driven solutions for connected vehicle networks, V2X-LLM pioneers the use of LLMs to process and interpret vast amounts of vehicular and infrastructural data in real-time. Built upon an advanced data pipeline system, the framework effectively addresses the complexities of connected vehicle corridors by dynamically acquiring high-fidelity vehicle record information, enabling near real-time estimation of traffic and environmental performance. LLMs are the most viable approach for this task due to their multi-tasking capabilities, ability to handle complex and heterogeneous data sources, and strong contextual reasoning. Unlike traditional rule-based or machine learning models, LLMs can seamlessly integrate diverse data types—structured sensor outputs, textual reports, and dynamic traffic patterns—to provide a holistic and adaptive understanding of corridor operations. By leveraging these capabilities, V2X-LLM delivers dynamic, real-time feedback to facility users, local inhabitants, and system administrators, significantly improving decision-making, traffic management, and overall operational efficiency The major contributions of this research include:

\begin{itemize}
    \item Introduction of the V2X-LLM framework to improve V2X communication and data interpretation using LLMs for smart corridors.
    \item Design and implementation of an advanced data pipeline system tailored for integrating LLMs with V2X systems.
    \item Real-time data analysis and contextual comprehension to facilitate seamless navigation through connected vehicle corridors, thereby improving the overall trip experience.
    \item Execution of four key tasks: Scenario Explanation, providing detailed narratives of traffic conditions; V2X Data Description, offering precise accounts of vehicle and infrastructure status; State Prediction, forecasting future traffic states; and Navigation Advisory, delivering optimized routing instructions. These tasks collectively enable enhanced traffic management and safety, as evidenced by comprehensive experimental results demonstrating the framework's effectiveness in improving V2X integration and understanding.
\end{itemize}

\section{V2X-LLM Framework}
\label{sec:framework}

\subsection{Framework Architecture}

The V2X-LLM framework is a structured system designed to integrate V2X data from connected vehicle corridors with advanced reasoning capabilities powered by a LLM. The architecture consists of multiple interconnected layers, each crucial for processing, interpretation, and synthesis of V2X information to support intelligent transportation systems.

\begin{figure}[!ht]
  \centering
  \includegraphics[width=0.47\textwidth]{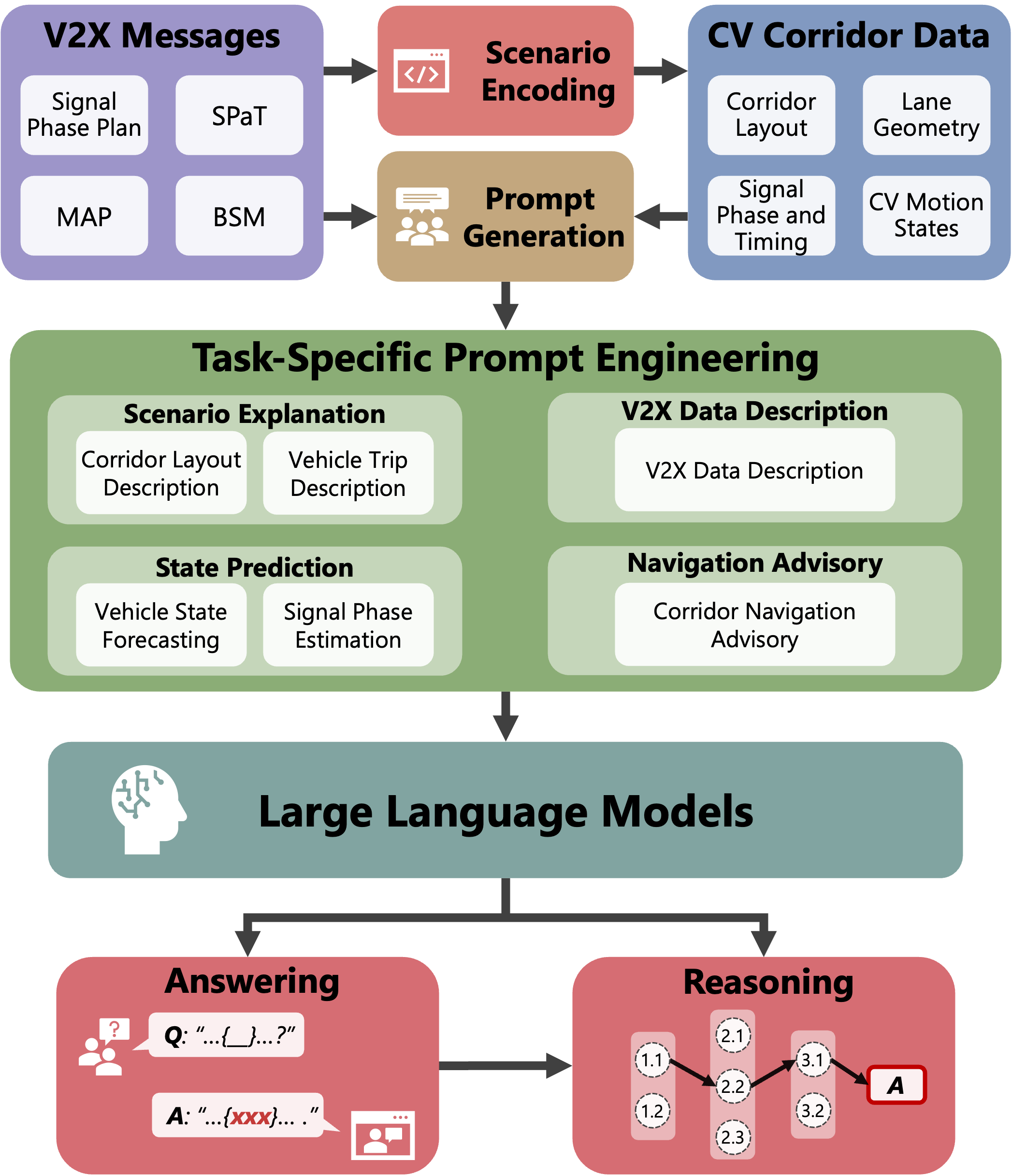}
  \caption{V2X-LLM Framework Architecture}\label{fig:framework}
\end{figure}

As shown in Figure~\ref{fig:framework}, the initial layer handles the collection and management of various V2X data inputs acquired from the data pipeline system~\cite{wu2023development, li2023does, wu2024digital}, including the coordinated signal phase plan design by the City of Madison, and SPaT data for historical and real-time signal state information for all the intersections along this connected vehicle corridor. Additionally, MAP data is incorporated to describe geometric layout of an intersection with lane-level details, while BSM data provide dynamic and static vehicle status. 

Next, the Scenario Encoding phase structures and formats raw V2X data into a coherent and standardized representation. This process involves extracting essential features from the data pipeline and encoding them into defined categories, such as Corridor Layout, which characterizes the spatial configuration of the CV corridor, and Lane Geometry, which specifies lane attributes. SPaT data is processed to provide real-time signal state assessments and predictive insights, while CV Motion States capture dynamic elements such as vehicle speed, position, and heading. 

The Prompt Generation phase translates encoded V2X data into structured prompts that serve as inputs for the LLM. This phase encompasses several key components, including Scenario Explanation, which generates descriptive narratives of the traffic environment, and V2X Data Description, which provides detailed insights into vehicle and infrastructure statuses. Additionally, the State Prediction module forecasts future vehicle states and estimates signal phase transitions, supporting proactive traffic management. The Navigation Advisory module generates recommendations for optimal routing and maneuvering strategies, aiding real-time decision-making in dynamic traffic conditions.

At the core of the framework, the LLM processes these structured prompts to generate meaningful outputs. The system produces two primary categories of responses: Answering and Reasoning. The Answering component delivers direct responses to specific queries derived from V2X data, ensuring timely and context-aware information retrieval. In contrast, the Reasoning component performs in-depth analysis by leveraging the LLM's capacity to infer patterns, establish contextual relationships, and synthesize complex insights. This dual-processing capability enhances situational awareness and provides stakeholders with a comprehensive understanding of evolving traffic conditions.

\subsection{Scenario Encoding}
The Scenario Encoding process serves as a fundamental element within the V2X-LLM framework, responsible for synthesizing and structuring diverse V2X data sources into a coherent format that enables systematic analysis and informed decision-making. This transformation is achieved through the Extractor module, which operates within the data pipeline~\cite{wu2023development, wu2024digital} and is designed to systematically process V2X messages. The module extracts essential data attributes from multiple sources and organizes them into a structured format that facilitates seamless integration with the LLM Reasoning Module for advanced traffic analysis and inference.

\begin{figure*}[!ht]
  \centering
  \includegraphics[width=0.9\textwidth]{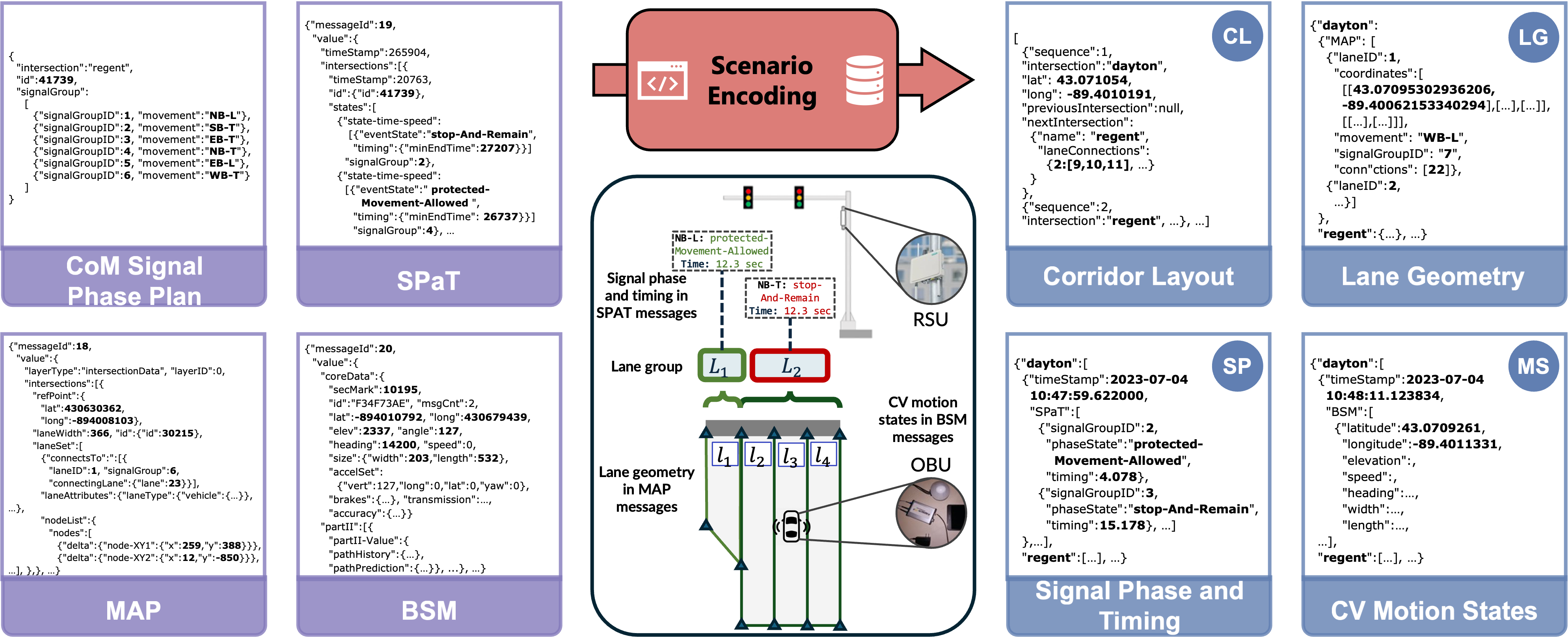}
  \caption{Scenario Encoding} \label{fig:v2x_data_scenario_encoding}
\end{figure*}

Figure~\ref{fig:v2x_data_scenario_encoding} illustrates the example data format of the raw V2X messages and the encoded CV corridor data. The Scenario Encoding includes several key components:

\begin{itemize}
    \item \textbf{Corridor Layout}: This component captures the spatial organization of the CV corridor, including parameters such as intersection identifiers, reference coordinates, and lane connectivity between intersections. By encoding this information, the system establishes a structural representation of the corridor.

    \item \textbf{Lane Geometry}: Extracted from MAP messages, this component provides a detailed characterization of lane attributes, including lane IDs, polygonal boundaries defining lane structures, permissible movements, signal group assignments, and inter-lane connections. This information is essential for understanding the specific configurations that regulate vehicular movements.

    \item \textbf{Signal Phase and Timing}: Extracted from SPaT messages, this includes data on signal group IDs, current phase states, and the remaining time for each phase. Accurate interpretation of this data supports the synchronization of vehicle movements with signal operations.

    \item \textbf{CV Motion States}: Derived from BSMs, this component captures real-time vehicular dynamics, including latitude, longitude, elevation, speed, heading, and vehicle dimensions. By maintaining a continuous stream of motion state data, the system generates a granular representation of vehicle's status and trajectory within the corridor.

\end{itemize}

The structured data output generated through Scenario Encoding serves as the foundation for prompt generation, enabling the LLM to analyze evolving traffic conditions, predict future vehicle states, and generate navigation advisories. The transformation of raw V2X data into structured representations enhances the interpretability of complex traffic environments and supports intelligent decision-making within the V2X-LLM framework.

\subsection{Role and Context of V2X-LLM}
The V2X-LLM framework functions as an AI-powered expert assistant within the connected and automated driving ecosystem, facilitating seamless V2X communication between CVs and RSUs deployed at multiple intersections along an urban arterial corridor. The primary objective of the V2X-LLM is to coordinate vehicle movements, enhancing situational awareness and facilitating efficient and safe navigation through complex traffic scenarios. The following box presents the role and context of the LLM framework.

\vspace{-1pt}
\begin{mdframed}[style=LLMFrame,linecolor=ForestGreen]
    % \small
    \scriptsize
    \textbf{LLM role}: You are an expert AI assistant of a connected and automated driving system, enabling V2X communication between connected vehicles and RSUs located at multiple intersections. Your role includes coordinating connected vehicles as they navigate through an urban arterial corridor.\\
    \textbf{Context}: 
    \begin{itemize}
        \item \textbf{Corridor Layout}: the fundamental details about each intersection equipped with an RSU within the corridor, including the intersection names and their reference coordinates.
        \item \textbf{Lane Geometry}: the lane geometry information extracted from the raw MAP messages, including lane IDs, the coordinates of polygons defining each lane, lane movement, signal group affiliations, and the IDs of connected lanes.
        \item \textbf{Signal Phase and Timing}: the signal phase information extracted from the raw MAP messages, including the signal group ID, the current phase state, and the corresponding remaining time at each timestamp.
        \item \textbf{CV Motion States}: the vehicle motion information extracted from the raw BSM messages, including the information of the latitude, longitude, elevation, speed, heading at each timestamp, and vehicle dimensions (width and height).
    \end{itemize}
\end{mdframed}
\vspace{-1pt}

The key to the framework's functionality is the detailed contextual data it utilizes. This includes critical spatial layout about the corridor's intersections and their associated RSUs, as well as comprehensive lane geometry data. This information, extracted from MAP messages, includes lane IDs, signal group affiliations, and other relevant characteristics that help in understanding traffic patterns and optimizing flow. Additionally, V2X-LLM incorporates signal phase and timing data, along with dynamic vehicle motion data, to provide a holistic view of the traffic environment. These components collectively enable the system to manage and streamline vehicle movements, enhancing overall traffic efficiency and safety.

\subsection{Prompt Design}

\subsubsection{Scenario Explanation}
The Scenario Explanation prompt is designed to generate detailed narratives about the traffic conditions experienced by connected vehicles. It instructs the V2X-LLM to analyze multiple data elements, including signal phase states, lane configurations, and vehicle movement patterns, to construct a comprehensive description of the traffic environment. The template guides the system to describe the corridor layout, detailing the number of RSU-equipped intersections, their directional alignment, and spatial distribution. Additionally, it provides insights into a vehicle's trip, specifying departure and arrival locations, the travel path, distances between intersections, and total travel time. 

\vspace{-1pt}
\begin{mdframed}[style=LLMFrame]
    % \small
    \scriptsize
    \textbf{Prompt}: Provide detailed narratives describing the traffic scenarios encountered by connected vehicles. This includes interpreting the data related to signal phases, lane configurations, and vehicle states, and explaining the interactions and potential outcomes of these elements within the traffic system. Explain the reasoning process.\\
    \textbf{Template}: 
    
    \begin{itemize}
        \item \textbf{Corridor Layout Description}: The connected vehicle corridor contains \textcolor{color1}{\{NUMBER\}} intersections with RSU installed. The corridor is aligned in \textcolor{color1}{\{DIRECTION\}} direction. These intersections are: \textcolor{color1}{\{INTERSECTION\_NAMES\}}. The distance between these intersections are \textcolor{color1}{\{DISTANCES\}}.
        \item \textbf{Vehicle Trip Description}: The connected vehicle started trip from \textcolor{color1}{\{INTERSECTION\_NAME\}} at \textcolor{color1}{\{TIMESTAMP\}} and travels along \textcolor{color1}{\{INTERSECTION\_NAMES\}}, and finally arrives at \textcolor{color1}{\{INTERSECTION\_NAME\}} at \textcolor{color1}{\{TIMESTAMP\}}. The vehicles travels \textcolor{color1}{\{DISTANCE\}} with \textcolor{color1}{\{DURATION\}}.
    \end{itemize}
\end{mdframed}
\vspace{-1pt}

\subsubsection{V2X Data Description}
The V2X Data Description prompt is structured to extract and present key aspects of V2X data, including lane geometry, signal phase states, and vehicle motion attributes. This prompt guides the V2X-LLM in identifying and specifying a vehicle's location based on lane ID and intersection placement, as well as its speed and direction of movement. Additionally, it incorporates signal phase and timing data, which are essential for evaluating a vehicle's current status and available maneuvering options. The structured prompt ensures that traffic conditions are represented with precision, allowing for an accurate assessment of vehicle positioning within the corridor.

\vspace{-1pt}
\begin{mdframed}[style=LLMFrame]
    % \small
    \scriptsize
    \textbf{Prompt}: Provide a comprehensive description of extracted V2X data including the lane geometry data, the signal phase and timing data, and CV motion states data. This involves describing the current location of the vehicle, the current lane the vehicle is located in at the intersection, the potential movement, and the corresponding signal phase and timing of the lane. Provide your explanations. Explain the reasoning process.\\
    \textbf{Template}: At the current timestamp \textcolor{color1}{\{TIMESTAMP\}}, the vehicle is currently in lane \textcolor{color1}{\{LANE\_ID\}} at intersection \textcolor{color1}{\{INTERSECTION\_NAME\}}, traveling at a speed of \textcolor{color1}{\{SPEED\}} miles per hour. The movement for this lane is \textcolor{color1}{\{MOVEMENT\}}. The signal phase for this lane is \textcolor{color1}{\{PHASE\_STATE\}}, with \textcolor{color1}{\{TIMING\}} seconds remaining.
\end{mdframed}
\vspace{-1pt}

\subsubsection{State Prediction}
The State Prediction prompt is designed to enable V2X-LLM to forecast future traffic states and vehicle dynamics based on historical and real-time data. The structured prompt template directs the model to predict a vehicle's trajectory over a five-second window and to estimate signal phase transitions at intersections. The system provides a breakdown of the current signal phase, remaining time, and the anticipated next phase. This forecasting capability assists in anticipating traffic patterns, identifying congestion points, and optimizing signal timing strategies, ultimately contributing to improved traffic flow management.

\vspace{-1pt}
\begin{mdframed}[style=LLMFrame]
    % \small
    \scriptsize
    \textbf{Prompt}: Analyze current and historical data to predict future states of traffic and vehicle dynamics. This includes forecasting traffic flow patterns, identifying potential congestion points, and predicting the timing and sequence of signal changes to optimize traffic movement. Explain the reasoning process.\\
    \textbf{Template}: 
    \begin{itemize}
        \item \textbf{Vehicle State Forecasting}: The vehicle is currently at \textcolor{color1}{\{(x0,x1)\}}.The predicted future 5-second driving trajectory of the vehicle will be \textcolor{color1}{\{[(x1,y1), (x2,y2), (x3,y3), (x4,y4), (x5,y5)]\}} (one per second). 
        \item \textbf{Signal Phase Estimation}: At the timestamp \textcolor{color1}{\{TIMESTAMP\}}, the signal phase and timing of intersection \textcolor{color1}{\{INTERSECTION\_NAME\}} are as follows:
        \begin{itemize}
            \item Lane \textcolor{color1}{\{LANE\_ID\}}: phase \textcolor{color1}{\{SIGNAL\_PHASE\}} with \textcolor{color1}{\{REMAINING\_TIME\}} remaining. The next phase and the anticipated time is \textcolor{color1}{\{NEXT\_SIGNAL\_PHASE\}} at \textcolor{color1}{\{NEXT\_TIMESTAMP\}}.
        \end{itemize}
    \end{itemize}
\end{mdframed}
\vspace{-1pt}

\subsubsection{Navigation Advisory}
The Navigation Advisory prompt is designed to generate real-time routing instructions for connected vehicles by taking into account traffic conditions, signal timings, and vehicle movements. This prompt instructs the V2X-LLM to assess the vehicle's current lane, associated signal group, and allowable maneuvers, before generating precise recommendations for lane and intersection transitions. The system also estimates travel times and advises on the most efficient and safe routes. By leveraging live traffic data, this structured advisory approach enhances vehicle coordination and contributes to efficient urban navigation.

\vspace{-1pt}
\begin{mdframed}[style=LLMFrame]
    % \small
    \scriptsize
    \textbf{Prompt}: Provide guidance and recommendations to connected vehicles on optimal navigation routes. This involves considering current traffic conditions, signal timings, and vehicle priorities to suggest the most efficient and safest routes through the urban corridor. Explain the reasoning process.\\
    \textbf{Template}: At current timestamp \textcolor{color1}{\{TIMESTAMP\}}, the connected vehicle is currently in lane \textcolor{color1}{\{LANE\_ID\}} at intersection \textcolor{color1}{\{INTERSECTION\_NAME\}}, with signal group \textcolor{color1}{\{SIGNAL\_GROUP\_ID\}} and movement \textcolor{color1}{\{MOVEMENT\}}. The signal phase is \textcolor{color1}{\{SIGNAL\_PHASE\}} and the remaining time is \textcolor{color1}{\{REMAINING\_TIME\}}. The vehicle should travel to \textcolor{color1}{\{LANE\_ID\}} inside the current intersection at current timestamp \textcolor{color1}{\{TIMESTAMP\}}, and travel to \textcolor{color1}{\{LANE\_ID\}} of the next intersection, the estimated travel time is \textcolor{color1}{\{ESTIMATED\_TRAVEL\_TIME\}}. At timestamp \textcolor{color1}{\{TIMESTAMP\}},(Starting from the next intersection, do the same for each one). And finally at timestamp \textcolor{color1}{\{TIMESTAMP\}}, the vehicle arrives in lane \textcolor{color1}{\{LANE\_ID\}} at intersection \textcolor{color1}{\{INTERSECTION\_NAME\}}.
\end{mdframed}
\vspace{-1pt}

\section{Experiment Design and Results}
\label{sec:experiment}

\subsection{Experiment Setup}
This research involved a series of field experiments~\cite{li2023does} conducted between April and July 2023 along the connected vehicle corridor on Park Street in Madison, USA. This arterial route, which connects University Avenue to downtown Madison, had a total of 15 Roadside Units RSUs installed. For our experiments, six of these RSUs were selected. The road section we primarily conduct our experiment consists of these six RSUs, spanning approximately 1.3 miles of Park Street. A Data Pipeline system was employed to retrieve and decode messages, ensuring efficient data extraction. To facilitate communication with the RSUs, a Cohda MK6C OBU equipped with antennas for Global Navigation Satellite System (GNSS) was mounted on the vehicle. Additionally, a laptop connected via Ethernet was used to receive data transmitted from the OBU.

For the LLM Reasoning Module, we adopt ChatGPT-4, a large language model, which is a large multi-model that can analyze text and picture inputs and produce text~\cite{achiam2023gpt}, for conducting explanation and reasoning in our scenarios.

\subsection{Results and Discussion}
\subsubsection{Scenario Explanation}
\begin{figure}[!t]
  \centering
  \includegraphics[width=0.42\textwidth]{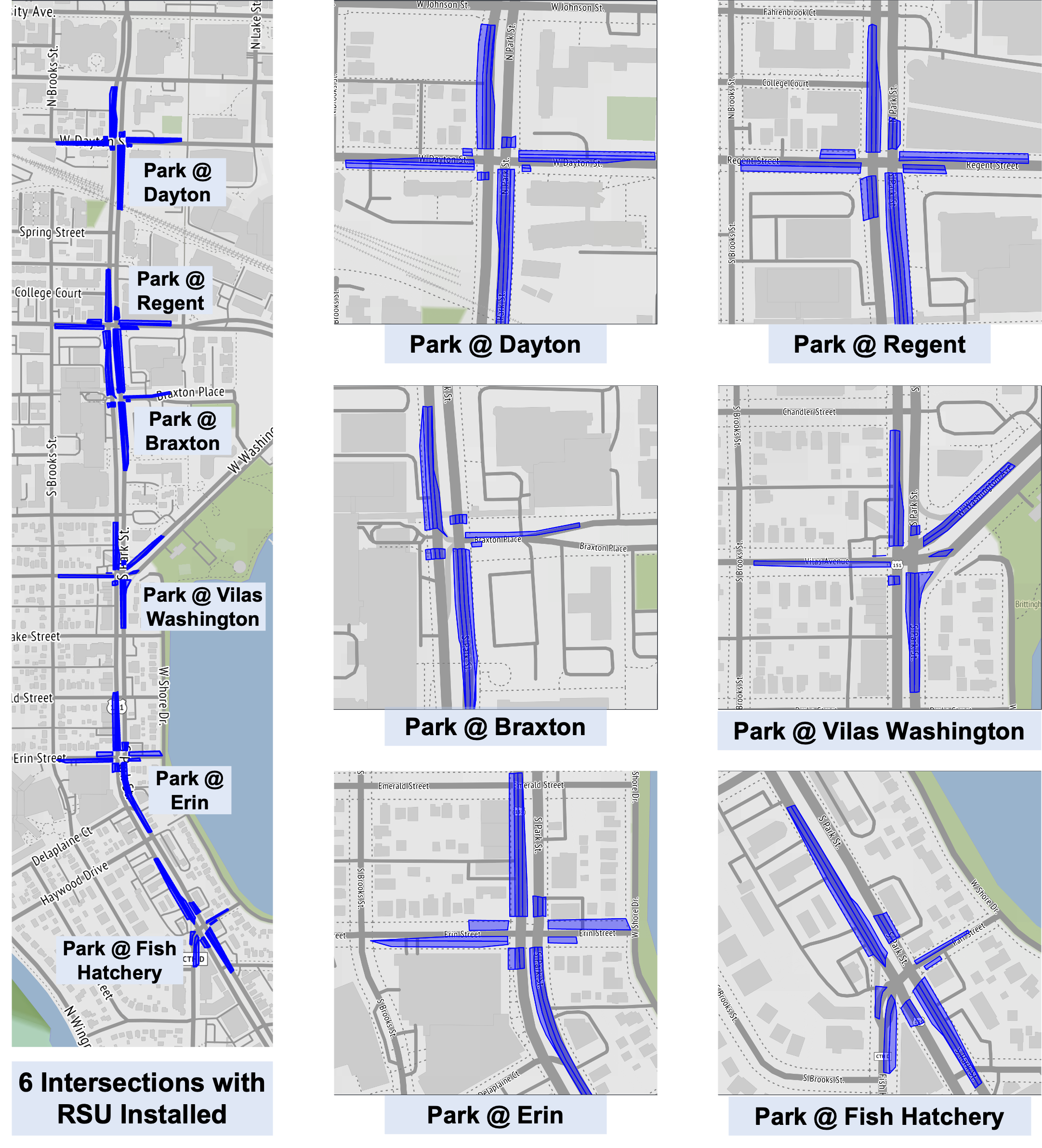}
  \caption{Visualization of Experiment 1: Scenario Explanation}\label{fig:v2x_scene_expl}
\end{figure}

The results of the Scenario Explanation task validate the V2X-LLM framework's ability to accurately interpret and describe the CV corridor layout. The system successfully extracts key spatial details, including the number of RSU-equipped intersections, their directional alignment, and the precise distances between them. The experiment's structured output, highlighted in yellow within the response boxes, follows the predefined template. Figure~\ref{fig:v2x_scene_expl} provides a visual representation of the intersections. The system identifies six intersections, all equipped with RSUs, running in a North-South direction across a total corridor length of 1.08 miles. Additionally, the framework calculates the exact distances between consecutive intersections, such as 0.24 miles from Park St @ Dayton to Park St @ Regent, 0.09 miles from Park St @ Regent to Park St @ Braxton, and similar measurements for other segments. 

Beyond extracting spatial attributes, V2X-LLM establishes a logical explanation and reasoning process. The system analyzes how intersection spacing influences vehicle movement and travel times, emphasizing its role in signal phase coordination and congestion management. By integrating BSM and SPaT data, the framework enhances its reasoning capabilities, using real-time data to assess vehicle interactions, optimize signal timings, and improve overall traffic efficiency. 

\vspace{-1pt}
\begin{mdframed}[style=LLMFrame, linecolor=BrickRed]
    % \small
    \scriptsize
    \textbf{Experiment}: Scenario Explanation: Corridor Layout Description\\
    \textbf{Answer}: The connected vehicle corridor contains \hl{six} intersections with RSU installed. The corridor is aligned in a general \hl{North-South} direction. These intersections are: \hl{Park St @ Dayton}, \hl{Park St @ Regent}, \hl{Park St @ Braxton}, \hl{Park St @ Vilas Washington}, \hl{Park St @ Erin}, and \hl{Park St @ Fish Hatchery}. The length of the corridor section is approximately \hl{1.08 miles}. The distances between each pair of consecutive intersections, e.g. from Park St @ Dayton to Park St @ Regent is approximately \hl{0.24 miles}, from Park St @ Regent to Park St @ Braxton is approximately \hl{0.09 miles}, from Park St @ Braxton to Park St @ Vilas Washington is approximately \hl{0.21 miles}, from Park St @ Vilas Washington to Park St @ Erin is approximately \hl{0.19 miles}, and from Park St @ Erin to Park St @ Fish Hatchery is approximately \hl{0.35 miles}.\\
    \textbf{Explanation and Reasoning}: The corridor consists of six intersections equipped with RSUs, providing a communication network for connected vehicles. The corridor runs in a North-South direction, facilitating vehicle movement through urban areas. The distances between intersections vary, influencing travel times and traffic flow. The data from the intersections and the vehicle's position within specific lanes are used to determine the appropriate signal phase and timing information, ensuring efficient traffic management and safety. 
\end{mdframed}
\vspace{-1pt}

\subsubsection{V2X Data Description}
\begin{figure*}[!ht]
  \centering
  \includegraphics[width=0.88\textwidth]{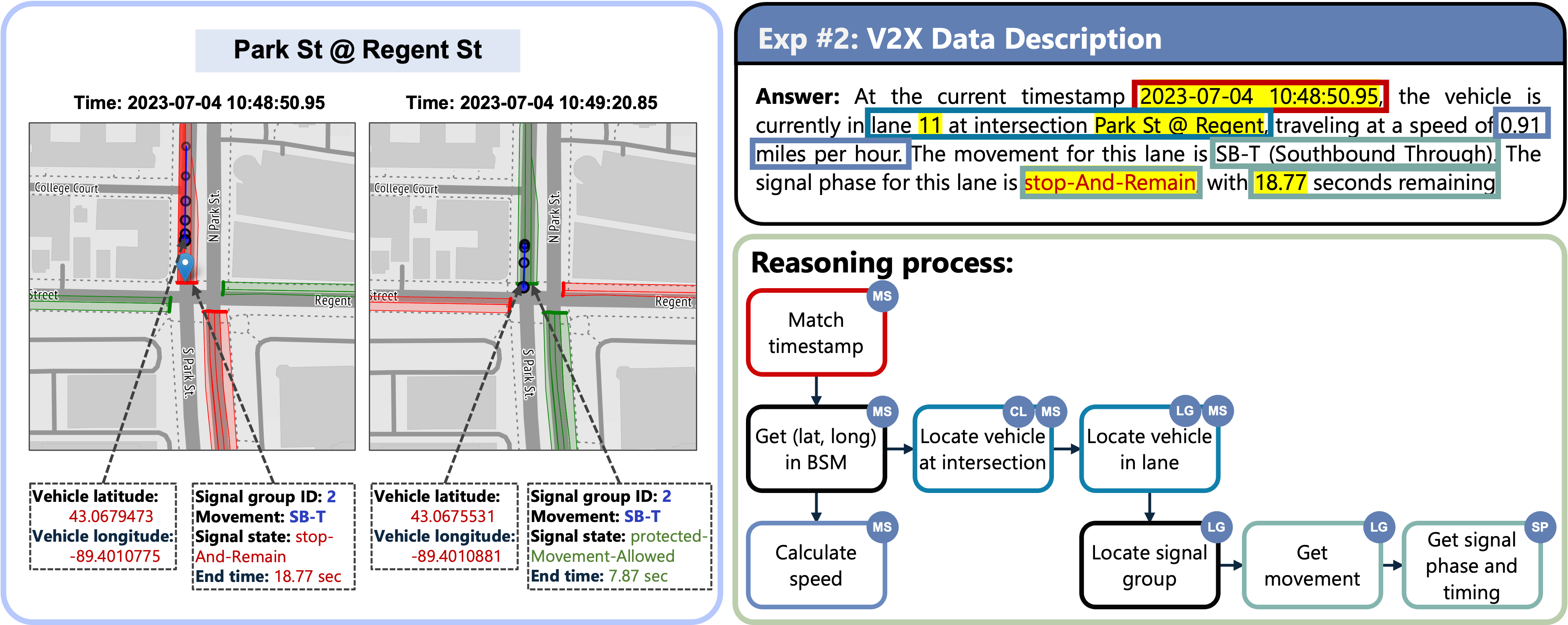}
  \caption{Experiment 2: V2X Data Description}\label{fig:v2x_data_descr}
\end{figure*}

The results of the V2X Data Description task illustrate the V2X-LLM framework's ability to accurately determine vehicle position, movement characteristics, and signal phase states based on real-time V2X data. The system effectively extracts and interprets the vehicle's exact lane location, speed, movement classification, and corresponding traffic signal phase. As demonstrated in Figure~\ref{fig:v2x_data_descr}, the experiment confirms that at timestamp 2023-07-04 10:48:50.95, the vehicle is identified in lane 11 at the ``Park St @ Regent'' intersection, traveling at a speed of 0.91 mph. The designated movement for this lane is Southbound Through (SB-T), and the signal phase is recorded as ``stop-And-Remain'', with 18.77 seconds remaining before transitioning.

\vspace{-1pt}
\begin{mdframed}[style=LLMFrame, linecolor=BrickRed]
    % \small
    \scriptsize
    \textbf{Experiment}: V2X Data Description\\
    \textbf{Answer}: At the current timestamp \hl{2023-07-04 10:48:50.95}, the vehicle is currently in lane \hl{11} at intersection \hl{Park St @ Regent}, traveling at a speed of \hl{0.91} miles per hour. The movement for this lane is \hl{SB-T (Southbound Through)}. The signal phase for this lane is \hl{stop-And-Remain}, with \hl{18.77} seconds remaining.\\
    \textbf{Explanation}: To determine the vehicle's location and traffic signal state, the system first matches the timestamp from the dataset to ensure it processes the most relevant BSM and SPaT data. Next, it retrieves the vehicle's latitude and longitude coordinates from the BSM data and associates these coordinates with the nearest intersection using geospatial mapping. After identifying the intersection as ``Park St @ Regent'', the system calculates the geodesic distance between the vehicle's position and lane boundaries, determining that the vehicle is in lane 11. The system then derives the vehicle's speed by analyzing positional changes over time, confirming a movement speed of 0.91 mph.
\end{mdframed}
\vspace{-1pt}

The reasoning process for these results follows a structured approach, integrating BSM and SPaT data to determine vehicle position and signal status. As outlined in the dialogue box, the system first matches the timestamp to ensure synchronization with the most relevant dataset. Using BSM data, it retrieves the vehicle's latitude and longitude coordinates and associates them with the closest intersection through geospatial mapping. The specific lane identification is determined by calculating geodesic distances between the vehicle's position and predefined lane boundaries, pinpointing lane 11 as the most probable location. Additionally, the system computes vehicle speed based on positional displacement over time, verifying the 0.91 mph movement. To establish traffic signal state information, the SPaT data is processed to identify the signal group controlling lane 11. The system retrieves the current signal phase, confirming that the lane is under a ``stop-And-Remain'' phase with 18.77 seconds left before transitioning. By integrating spatial and signal timing data, the V2X-LLM ensures accurate real-time traffic assessment, supporting improved vehicle coordination and enhanced intersection safety.

\subsubsection{State Prediction}

\begin{figure*}[!ht]
  \centering
  \includegraphics[width=0.88\textwidth]{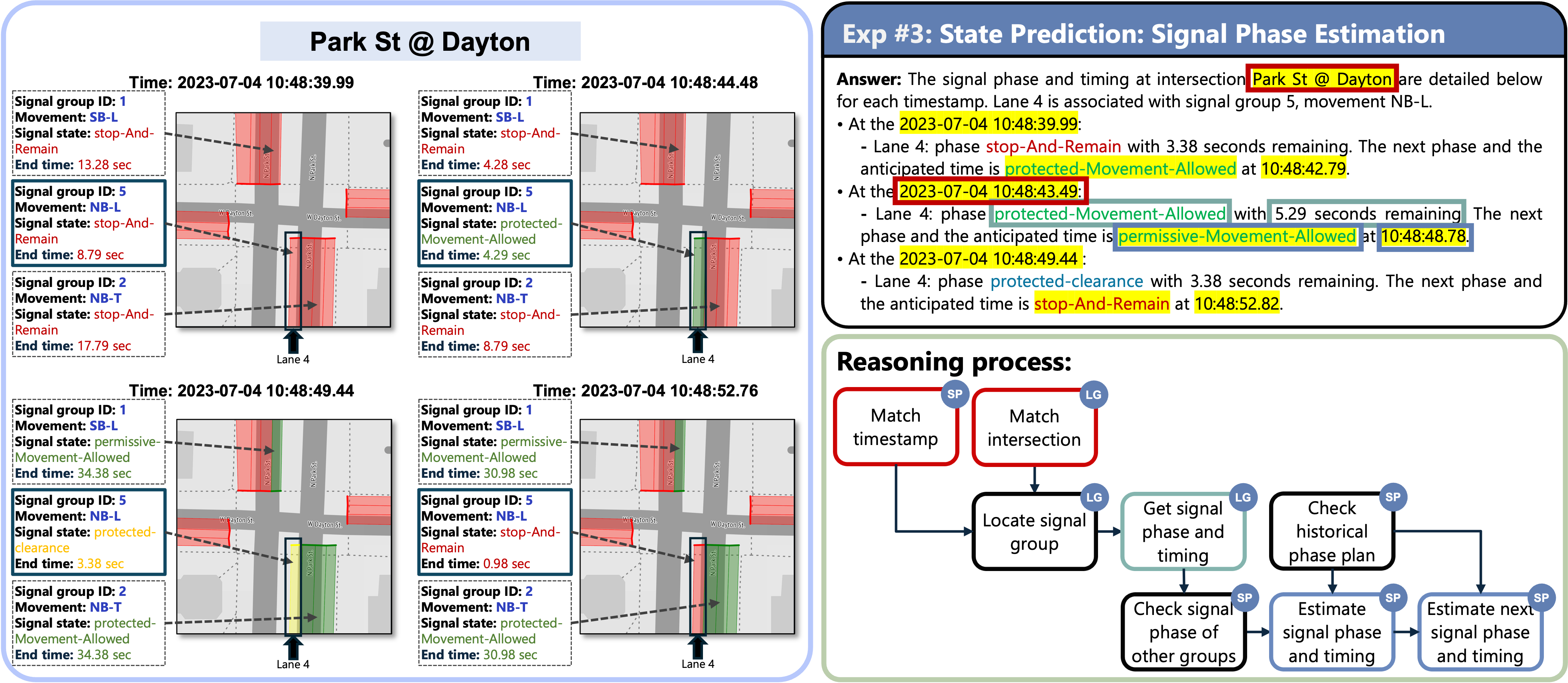}
  \caption{Experiment 3: State Prediction - Signal Phase Estimation}\label{fig:state_prediction}
\end{figure*}

The State Prediction task evaluates the V2X-LLM framework's capability to forecast future traffic signal states and phase transitions based on SPaT data. The experiment focuses on estimating phase changes at the "Park St @ Dayton" intersection, as shown in Figure~\ref{fig:state_prediction}. The system predicts transitions for lane 4 (NB-L), lane 6 (NB-T), and lane 11 (SB-L), each assigned to a distinct signal group. The results demonstrate that V2X-LLM effectively anticipates phase shifts and remaining times, such as the transition from ``stop-And-Remain'' to ``protected-Movement-Allowed'', followed by ``permissive-Movement-Allowed'', ensuring accurate synchronization with standard traffic signal cycles.

\vspace{-1pt}
\begin{mdframed}[style=LLMFrame, linecolor=BrickRed]
    % \small
    \scriptsize
    \textbf{Experiment}: State Prediction: Signal Phase Estimation\\
    \textbf{Answer}: \\
    The signal phase and timing at intersection \hl{Park St @ Dayton} are detailed below for each timestamp. Lane \hl{11} is associated with signal Group \hl{1}, movement \hl{SB-L}. Lane \hl{6} is associated with signal Group \hl{2}, movement \hl{NB-T}. Lane \hl{4} is associated with signal Group \hl{5}, movement \hl{NB-L}.\\
    At the timestamp \hl{2023-07-04 10:48:39.99}, the signal phase and timing are as follows:
    \begin{itemize}
    \item Lane 11: phase stop-And-Remain with 2.79 seconds remaining. The next phase and the anticipated time is \hl{permissive-Movement-Allowed} at \hl{10:48:42.79}.
    \item Lane 6: phase stop-And-Remain with 2.79 seconds remaining. The next phase and the anticipated time is \hl{permissive-Movement-Allowed} at \hl{10:48:42.79}.
    \item Lane 4: phase stop-And-Remain with 2.79 seconds remaining. The next phase and the anticipated time is \hl{protected-Movement-Allowed} at \hl{10:48:42.79}.
    \end{itemize}
    % At the timestamp \hl{2023-07-04 10:48:43.49}, the signal phase and timing are as follows:
    % \begin{itemize}
    % \item Lane 11: phase stop-And-Remain with -0.01 seconds remaining. The next phase and the anticipated time is \hl{permissive-Movement-Allowed} at \hl{10:48:43.48}.
    % \item Lane 6: phase stop-And-Remain with -0.01 seconds remaining. The next phase and the anticipated time is \hl{permissive-Movement-Allowed} at \hl{10:48:43.48}.
    % \item Lane 4: phase protected-Movement-Allowed with 5.29 seconds remaining. The next phase and the anticipated time is \hl{permissive-Movement-Allowed} at \hl{10:48:48.78}.
    % \end{itemize}
    At the timestamp \hl{2023-07-04 10:48:49.44}, the signal phase and timing are as follows:
    \begin{itemize}
    \item Lane 11: phase permissive-Movement-Allowed with 34.38 seconds remaining. The next phase and the anticipated time is \hl{stop-And-Remain} at \hl{10:49:23.82}.
    \item Lane 6: phase protected-Movement-Allowed with 34.38 seconds remaining. The next phase and the anticipated time is \hl{stop-And-Remain} at \hl{10:49:23.82}.
    \item Lane 4: phase protected-clearance with 3.38 seconds remaining. The next phase and the anticipated time is \hl{stop-And-Remain} at \hl{10:48:52.82}.
    \end{itemize}
    % At the timestamp \hl{2023-07-04 10:48:52.76}, the signal phase and timing are as follows:
    % \begin{itemize}
    % \item Lane 11: phase permissive-Movement-Allowed with 30.98 seconds remaining. The next phase and the anticipated time is \hl{stop-And-Remain} at \hl{10:49:23.74}.
    % \item Lane 6: phase protected-Movement-Allowed with 30.98 seconds remaining. The next phase and the anticipated time is \hl{stop-And-Remain} at \hl{10:49:23.74}.
    % \item Lane 4: phase stop-And-Remain with 0.98 seconds remaining. The next phase and the anticipated time is \hl{protected-Movement-Allowed} at \hl{10:48:53.74}.
    % \end{itemize}
    At the timestamp \hl{2023-07-04 10:48:57.93}, the signal phase and timing are as follows:
    \begin{itemize}
    \item Lane 11: phase permissive-Movement-Allowed with 25.79 seconds remaining. The next phase and the anticipated time is \hl{stop-And-Remain} at \hl{10:49:23.72}.
    \item Lane 6: phase protected-Movement-Allowed with 25.79 seconds remaining. The next phase and the anticipated time is \hl{stop-And-Remain} at \hl{10:49:23.72}.
    \item Lane 4: phase permissive-Movement-Allowed with 25.79 seconds remaining. The next phase and the anticipated time is \hl{stop-And-Remain} at \hl{10:49:23.72}.
    \end{itemize}
    \textbf{Explanation}: The system predicts future signal phase transitions by following a structured reasoning process. First, it matches the timestamp with the latest SPaT data to ensure synchronization with real-time traffic conditions. Next, it identifies the intersection, locating the corresponding signal group controlling lane 4. Once the system has determined the active signal phase and remaining time, it retrieves signal phase data from other groups to provide context for broader traffic interactions. To enhance accuracy, the system references historical phase plans, which outline typical phase sequences and durations. By checking previous signal transitions, the model estimates the current phase's expected duration before shifting to the next state. Finally, using the remaining time and observed phase cycles, the system calculates the precise timestamp for the next phase transition. For example, when lane 4 is in ``stop-And-Remain'' at 10:48:39.99 with 3.38 seconds left, the model projects the shift to ``protected-Movement-Allowed'' at 10:48:42.79. This process is systematically repeated across timestamps, ensuring that the system delivers accurate, real-time signal phase predictions for optimizing traffic flow and vehicle coordination.

\end{mdframed}
\vspace{-1pt}

The dialogue box details the system's signal phase estimations at multiple timestamps. At 10:48:39.99, the system predicts that lane 4 will remain in the ``stop-And-Remain'' phase for 3.38 seconds, transitioning to “protected-Movement-Allowed” at 10:48:42.79. Similarly, at 10:48:43.49, lane 4 is expected to transition again in 5.29 seconds, moving into ``permissive-Movement-Allowed'' at 10:48:48.78. At 10:48:49.44, the system estimates that lane 4 will enter a ``protected-clearance'' phase for 3.38 seconds, shifting back to ``stop-And-Remain'' at 10:48:52.82. These predictions validate the system's ability to track dynamic traffic signal behavior and provide real-time insights for optimizing vehicle movements and reducing congestion.

By aligning predicted transitions with real-time SPaT updates, the V2X-LLM framework enhances signal coordination and vehicle movement optimization. The system's ability to differentiate transitions across multiple lanes and movements further refines its traffic control predictions. For instance, while lane 4 transitions through multiple movement states, other lanes may retain their “protected-Movement-Allowed” status for extended durations. This differentiation highlights the framework's adaptability to complex intersection control strategies. Additionally, the detailed timing breakdown enables precise vehicle coordination, allowing connected vehicles to anticipate upcoming phase changes and adjust their speeds accordingly. The accurate prediction of phase transitions and clearance intervals helps mitigate unnecessary stops, reduce congestion, and improve overall traffic efficiency. By leveraging real-time SPaT data and historical phase analysis, the V2X-LLM contributes to dynamic and adaptive traffic signal control, reinforcing its potential for intelligent urban mobility solutions.

\subsubsection{Navigation Advisory}

\begin{figure*}[!ht]
  \centering
  \includegraphics[width=0.88\textwidth]{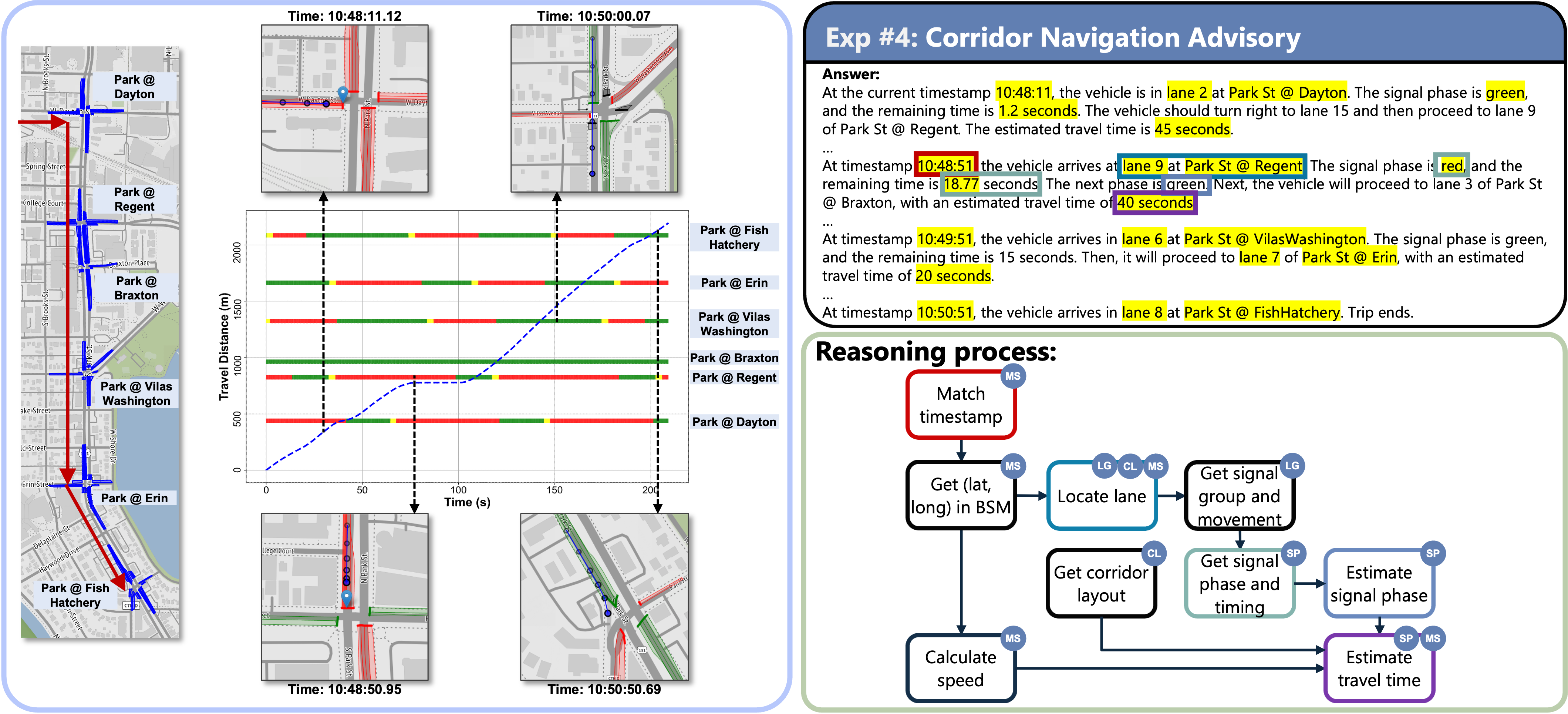}
  \caption{Experiment 4: Corridor Navigation Advisory}\label{fig:navigation_advisory}
\end{figure*}

The Navigation Advisory task assesses the V2X-LLM framework’s ability to generate structured, real-time routing guidance for vehicles traversing a connected urban corridor. The system integrates lane transition logic, signal phase compliance, and estimated travel times to optimize vehicle movement while ensuring adherence to traffic regulations. Figure~\ref{fig:navigation_advisory} presents the navigation advisory for a vehicle traveling from Park St @ Dayton to Park St @ Fish Hatchery, detailing each lane transition and corresponding signal phase interaction. The advisory continuously updates based on real-time SPaT data, aligning vehicle movement with green signal phases to minimize stops and improve traffic flow.

\vspace{-1pt}
\begin{mdframed}[style=LLMFrame, linecolor=BrickRed]
    % \small
    \scriptsize
    \textbf{Experiment}: Navigation Advisory\\
    \textbf{Answer}:\\
    At the current timestamp \hl{2023-07-04 10:48:11}, the connected vehicle is in lane \hl{2} at the intersection \hl{Park St @ Dayton}, with signal group \hl{4} and movement \hl{WB-T}. The signal phase is \hl{green}, and the remaining time is \hl{1 second}. The vehicle should travel to lane \hl{15} inside the current intersection at the current timestamp \hl{2023-07-04 10:48:11}, and then proceed to lane \hl{9} of the next intersection, \hl{Park St @ Regent}. The estimated travel time between intersections is \hl{40 seconds}. At timestamp \hl{2023-07-04 10:48:51}, the vehicle arrives at lane \hl{9} at \hl{Park St @ Regent}.\\
    Next, from lane \hl{9} at \hl{Park St @ Regent}, the vehicle will move to lane \hl{19} within the intersection. From there, it will proceed to lane \hl{3} of the next intersection, \hl{Park St @ Braxton}, with an estimated travel time of \hl{50 seconds}. At timestamp \hl{2023-07-04 10:49:41}, the vehicle arrives in lane \hl{3} at \hl{Park St @ Braxton}.\\
    Continuing, from lane \hl{3} at \hl{Park St @ Braxton}, the vehicle will navigate to lane \hl{10} within the intersection. Then, it will proceed to lane \hl{6} of the next intersection, \hl{Park St @ VilasWashington}, with an estimated travel time of \hl{10 seconds}. At timestamp \hl{2023-07-04 10:49:51}, the vehicle arrives in lane \hl{6} at \hl{Park St @ VilasWashington}.\\
    From lane \hl{6} at \hl{Park St @ VilasWashington}, the vehicle will transition to lane \hl{15} within the intersection. Then, it will proceed to lane \hl{7} of the next intersection, \hl{Park St @ Erin}, with an estimated travel time of \hl{25 seconds}. At timestamp \hl{2023-07-04 10:50:16}, the vehicle arrives in lane \hl{7} at \hl{Park St @ Erin}.\\
    Finally, from lane \hl{7} at \hl{Park St @ Erin}, the vehicle will move to lane \hl{16} within the intersection. It will then proceed to lane \hl{8} of the final intersection, \hl{Park St @ FishHatchery}, with an estimated travel time of \hl{35 seconds}. At timestamp \hl{2023-07-04 10:50:51}, the vehicle arrives in lane \hl{8} at \hl{Park St @ FishHatchery}.\\
    \textbf{Explanation}: The system generates the navigation advisory by first matching the timestamp with the latest available BSM data, ensuring the vehicle’s current location is accurately identified. It then extracts the latitude and longitude coordinates to determine the vehicle’s lane and intersection position. After identifying the current lane, the system retrieves the corridor layout to define possible lane transitions within and between intersections. It then associates the vehicle’s lane with the appropriate signal group and movement using SPaT data, allowing it to determine signal phases, remaining time, and movement restrictions at each step. The system then calculates estimated travel times based on intersection distances, signal phase durations, and historical traffic patterns. This estimation ensures that the vehicle receives realistic and actionable timing for each transition, minimizing unnecessary stops and maximizing efficiency. Finally, the advisory integrates all traffic, signal, and spatial data to generate a structured step-by-step route. The system ensures the vehicle moves optimally through the corridor, adjusting for real-time traffic conditions and signal phase changes to improve travel efficiency and reduce congestion. 
\end{mdframed}
\vspace{-1pt}

The dialogue box below outlines the generated navigation plan, including step-by-step lane transitions, estimated travel times, and real-time signal phase conditions. The V2X-LLM Navigation Advisory follows a structured reasoning process that ensures optimal vehicle movement through complex urban intersections. The process begins with timestamp matching to align the advisory with the latest BSM data, ensuring that vehicle positioning is accurate. This step is crucial for maintaining consistency with real-time traffic conditions. Once the vehicle’s location is established, the system retrieves the corridor layout and determines lane connectivity, ensuring that all suggested transitions comply with roadway infrastructure and signal group regulations. The signal phase evaluation further refines the advisory by integrating remaining phase durations and expected phase transitions, allowing the system to synchronize vehicle movement with green signals and minimize waiting times. The travel time estimation stage leverages historical congestion trends, real-time vehicle speeds, and intersection spacing to generate precise arrival time predictions. This ensures that connected vehicles can anticipate lane transitions well in advance, enhancing coordination with other road users and reducing congestion.

By structuring the advisory through a combination of spatial analysis, signal phase alignment, and predictive traffic modeling, the V2X-LLM framework delivers an adaptive and responsive navigation strategy that optimizes urban mobility.

\subsection{Performance Evaluation}
To evaluate the V2X-LLM framework, we conducted a comprehensive assessment across four core tasks: Scenario Explanation, V2X Data Description, State Prediction, and Navigation Advisory. A 35-minute field test was performed, during which 18,340 BSM messages and over 110,000 SPaT messages were collected. The data were structured into task-specific slices, with varying lengths based on the requirements of each task.

\begin{figure}[!t]
  \centering
  \includegraphics[width=0.46\textwidth]{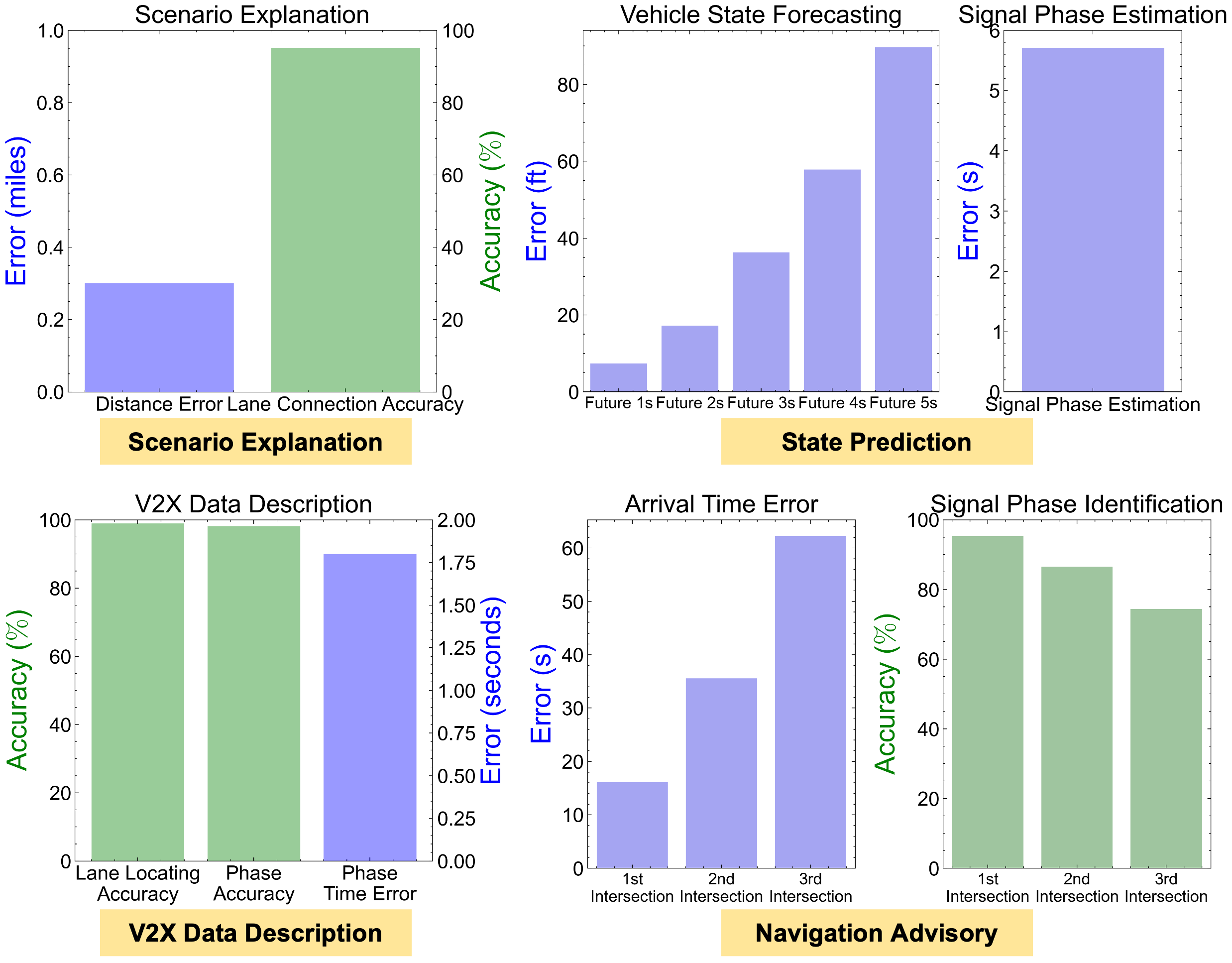}
  \caption{Performance Evaluation of the Four Experiments}\label{fig:eval}
\end{figure}

As illustrated in Figure~\ref{fig:eval}, the Scenario Explanation task demonstrated high lane connection accuracy ($95\%$), indicating reliable representation of lane configurations. However, the system exhibited a 0.3-mile distance error in estimating intersection distances, which may stem from map data limitations or sensor inaccuracies. For the V2X Data Description task, the system achieved $98.9\%$ lane identification accuracy and $98.1\%$ accuracy in traffic signal phase detection, validating its effectiveness in integrating and interpreting V2X data. Nonetheless, the phase time estimation exhibited an average error of 1.8 seconds, likely due to signal variability or transmission delays.

The State Prediction task showed an expected increase in Vehicle State Forecasting errors as the prediction horizon lengthened, ranging from 7.4 feet at 1 second to 89.6 feet at 5 seconds, reflecting the challenge of maintaining long-term prediction accuracy. Signal Phase Estimation demonstrated a 5.7-second error, which can be attributed to the complexity of predicting phase transitions under dynamic traffic conditions. For the Navigation Advisory task, the framework encountered increasing errors in arrival time predictions, measuring 16.1 seconds at the first intersection, 35.6 seconds at the second, and 62.2 seconds at the third, indicating a cumulative effect of uncertainties over longer travel distances. Additionally, the accuracy of signal phase identification declined from $95.2\%$ at the first intersection to $74.4\%$ at the third, underscoring the challenge of maintaining precision in long-range route planning and signal coordination.

\vspace{-2pt}
\section{Conclusions}

By incorporating advanced data pipelines and LLM-based analysis, the framework enhances real-time interpretation of traffic data, providing actionable insights into traffic conditions and infrastructure states. Designed to support intelligent transportation applications, the framework is structured around four core tasks: Scenario Explanation, which generates detailed descriptions of traffic situations; V2X Data Description, which captures and summarizes vehicle and infrastructure information; State Prediction, which forecasts future traffic states; and Navigation Advisory, which provides optimized routing recommendations. Together, these tasks improve the accuracy of traffic analysis, enhance safety, and facilitate more efficient traffic flow management.

The evaluation demonstrates V2X-LLM’s effectiveness in processing large-scale V2X data and generating accurate traffic insights. The framework shows high accuracy in lane identification and signal phase interpretation, effectively capturing real-time traffic conditions. However, forecasting challenges arise in long-term vehicle state predictions and signal phase estimations, where errors accumulate over time. Similarly, navigation accuracy declines as trip duration increases, affecting arrival time precision at later intersections. These findings highlight the framework’s strong data integration and reasoning capabilities, while also identifying areas for improvement in predictive modeling and real-time adjustments.

Despite its innovations, the V2X-LLM framework faces limitations, primarily related to data delays and the computational challenges of real-time processing. These issues can affect the system's responsiveness in dynamic traffic environments. Future research should focus on fine-tuning the LLM for improved accuracy in traffic scenarios and exploring hybrid AI approaches. Integrating Vision-Language Models (VLMs)~\cite{you2024v2x} and other neural networks could enhance predictive capabilities by combining visual perception with language-based reasoning. Additionally, optimizing computational efficiency and developing more adaptive processing techniques will be crucial for real-time applications.

\vspace{-2pt}
\section{Acknowledgements}
The Park Street Smart Corridor is being developed through a collaboration of the TOPS Lab, the City of Madison, Traffic and Parking Control Products and Solutions (TAPCO), and the Wisconsin Department of Transportation.  The ideas and views expressed in this paper are strictly those of the Traffic Operations and Safety (TOPS) Laboratory at the University of Wisconsin-Madison.

\vspace{-1pt}
\bibliographystyle{IEEEtranN}
\bibliography{ref}

\vspace*{-1em}
\begin{IEEEbiography}[{\includegraphics[width=1in,height=1.25in,clip,keepaspectratio]{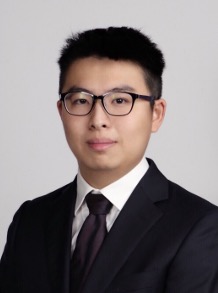}}]{Keshu Wu} is a postdoctoral research associate at Texas A\&M University. He receives his Ph.D. in Civil and Environmental Engineering from the University of Wisconsin-Madison in 2024. He also holds an M.S. degree in Civil and Environmental Engineering from Carnegie Mellon University in 2018 and an M.S. degree in Computer Sciences from the University of Wisconsin-Madison in 2022. He completed his B.S. in Civil Engineering at Southeast University in Nanjing, China in 2017. His research interests include the application and innovation of artificial intelligence and deep learning techniques in connected automated driving, intelligent transportation systems, and digital twin modeling and simulation.
\end{IEEEbiography}

\vspace*{-3em}
\begin{IEEEbiography}[{\includegraphics[width=1in,height=1.25in,clip,keepaspectratio]{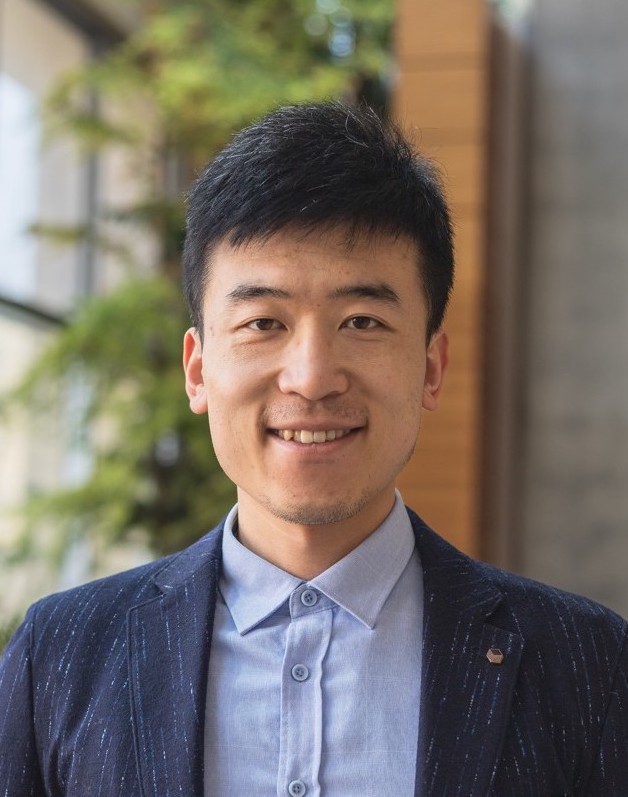}}]{Pei Li} is a Scientist in the Department of Civil and Environmental Engineering at the University of Wisconsin-Madison. He received his Ph.D. in Civil Engineering with a focus on Transportation Engineering from the University of Central Florida in 2021, after which he served as a Postdoctoral Research Fellow at the University of Michigan Transportation Research Institute. His research interests include transport safety, smart mobility, human factors, machine learning, connected and automated vehicles, and digital twins.
\end{IEEEbiography}

\vspace*{-3em}
\begin{IEEEbiography}
[{\includegraphics[width=1in,height=1.25in,clip,keepaspectratio]{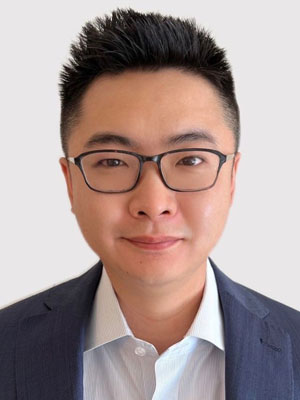}}]{Yang Zhou} received the Ph.D. degree in Civil and Environmental Engineering from University of Wisconsin Madison, WI, USA, in 2019, and the M.S. degree in Civil and Environmental Engineering from University of Illinois at Urbana-Champaign, Champaign, IL, USA, in 2015. He is currently an Assistant Professor in the Zachry Department of Civil and Environmental Engineering at Texas A\&M University. Before joining Texas A\&MUniversity, he was a postdoctoral researcher in Civil Engineering, University of Wisconsin Madison, WI, USA. He is currently a member in TRB traffic flow theory CAV subcommittee, network modeling CAV subcommittee, and American Society of Civil Engineering. His main research directions are connected automated vehicle robust control, interconnected system stability analysis, traffic big data analysis, and microscopic traffic flow modeling.
\end{IEEEbiography}

\vspace*{-3em}
\begin{IEEEbiography}[{\includegraphics[width=1in,height=1.25in,clip,keepaspectratio]{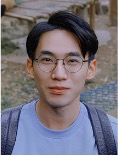}}]{Rui Gan} received his B.S. and M.S. degrees in Traffic Engineering from Southeast University in 2020 and 2022, respectively. He is currently pursuing a Ph.D. in Civil and Environmental Engineering at the University of Wisconsin-Madison. His research focuses on AI in intelligent connected autonomous vehicles and infrastructures, Vehicle motion prediction and planning and LLM-empowered autonomous driving systems.
\end{IEEEbiography}

\vspace*{-3em}
\begin{IEEEbiography}[{\includegraphics[width=1in,height=1.25in,clip,keepaspectratio]{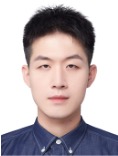}}]{Junwei You} received the M.S. degree in Civil and Environmental Engineering from Northwestern University in 2022. He is currently a Ph.D. student in Civil and Environmental Engineering at University of Wisconsin-Madison. His research interests are autonomous driving, foundation models, generative AI, and intelligent transportation systems.
\end{IEEEbiography}

\vspace*{-3em}
\begin{IEEEbiography}[{\includegraphics[width=1in,height=1.25in,clip,keepaspectratio]{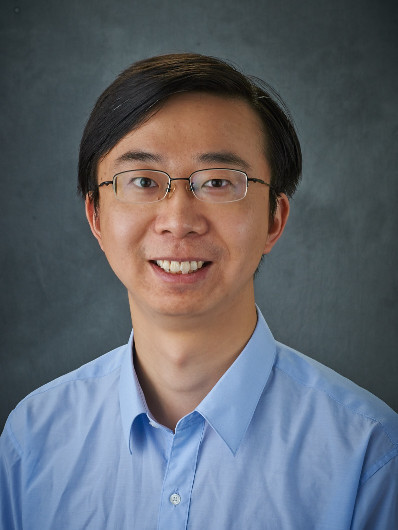}}]{Yang Cheng}
received the B.S. and M.S. degrees in automation from Tsinghua University, Beijing, China, in 2004 and 2006, respectively, and the Ph.D. degree in civil engineering from the University of Wisconsin–Madison in 2011. He is currently a scientist at the Wisconsin Traffic Operations and Safety (TOPS) Laboratory of the University of Wisconsin-Madison. His research areas include automated highway and driving systems, mobile traffic sensor modeling, large-scale transportation data management and analytics, and traffic operations and control.
\end{IEEEbiography}

\vspace*{-3em}
\begin{IEEEbiography}[{\includegraphics[width=1in,height=1.25in,clip,keepaspectratio]{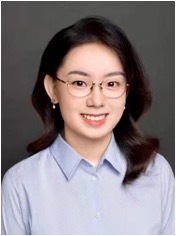}}]{Jingwen Zhu} is currently a Ph.D. student in Civil and Environmental Engineering at University of Wisconsin-Madison. Her research explores traffic safety and crash analysis, the application of machine learning in intelligent transportation systems, and the effects of different weather conditions on connected and automated vehicle highways.
\end{IEEEbiography}

\vspace*{-3em}
\begin{IEEEbiography}[{\includegraphics[width=1in,height=1.25in,clip,keepaspectratio]{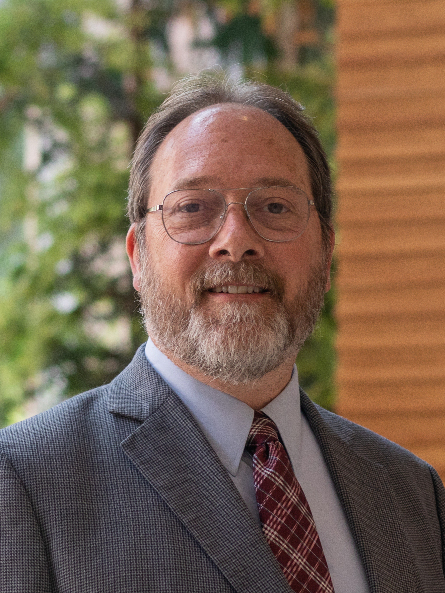}}]{Steven T. Parker}
is the Managing Director of the Wisconsin Traffic Operations and Safety (TOPS) Laboratory at the University of Wisconsin-Madison. He has led a range of research and development initiatives for the TOPS Lab across several core areas including transportation safety, work zone systems, traffic management systems, and connected and automated vehicle technologies. He received a Ph.D. in Computer Science from the University of Wisconsin-Madison. He is currently serving in his second term as the Chair of the Transportation Research Board (TRB) AED30 Information Systems and Technology Committee.
\end{IEEEbiography}

\vspace*{-3em}
\begin{IEEEbiography}[{\includegraphics[width=1in,height=1.25in,clip,keepaspectratio]{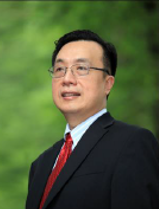}}]{Bin Ran}
is the Vilas Distinguished Achievement Professor and Director of ITS Program at the University of Wisconsin at Madison. Dr. Ran is an expert in dynamic transportation network models, traffic simulation and control, traffic information system, Internet of Mobility, Connected Automated Vehicle Highway (CAVH) System. He has led the development and deployment of various traffic information systems and the demonstration of CAVH systems. Dr. Ran is the author of two leading textbooks on dynamic traffic networks. He has co-authored more than 240 journal papers and more than 260 referenced papers at national and international conferences. He holds more than 20 patents of CAVH in the US and other countries. He is an associate editor of Journal of Intelligent Transportation Systems.
\end{IEEEbiography}

\vspace*{-3em}
\begin{IEEEbiography}[{\includegraphics[width=1in,height=1.25in,clip,keepaspectratio]{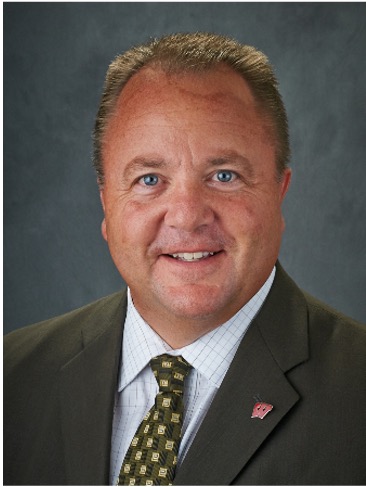}}]{David A. Noyce}
received his B.S. and M.S. degrees in Civil and Environmental Engineering from UW-Madison in 1984 and 1995, respectively, and received his Ph.D. degree in Civil (Transportation) Engineering from Texas A\&M University in 1999. He has authored more than 380 refereed scholarly papers, conference proceedings, research reports, and book chapters. He was elected Fellow in the American Society of Civil Engineers (ASCE) in 2017 and was President of ASCE's Transportation and Development Institute (T\&DI) in 2022. He works with the National Academy of Sciences and the Transportation Research Board (TRB), where he has chaired several National Cooperative Highway Research Program (NCHRP) project panels and has (and is currently) conducted NCHRP research.
\end{IEEEbiography}

\vspace*{-3em}
\begin{IEEEbiography}
[{\includegraphics[width=1in,height=1.25in,clip,keepaspectratio]{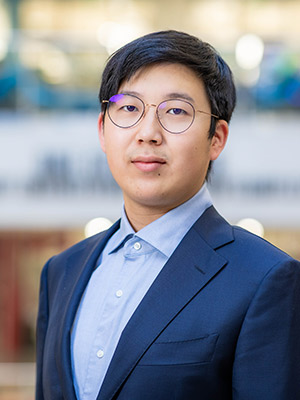}}]{Zhengzhong Tu} received the bachelor's and master's degrees from Fudan University, Shanghai, China, in 2016 and 2018, respectively, and the PhD degree from the University of Texas at Austin, Austin, TX, USA, advised by Professor Alan Bovik. He was a researcher with Google Research from 2022 to 2024. He is currently an assistant professor of computer Science with Texas A\&M University, College Station, TX. His research interests include generative AI, multimodal AI, and their applications in computational photography, autonomous driving, and robotics.
\end{IEEEbiography}

\end{document}